\pdfoutput=1

\documentclass[11pt]{article}

\usepackage{acl}

\usepackage{times}
\usepackage{latexsym}
\usepackage{amsmath}
\usepackage{enumitem}
\usepackage{hyperref}
\usepackage[T1]{fontenc}

\usepackage[utf8]{inputenc}

\usepackage{microtype}
\usepackage{inconsolata}

\usepackage{graphicx}

\title{Enhancing Domain-Specific Encoder Models with LLM-Generated Data: How to Leverage Ontologies, and How to Do Without Them}

\author{Marc Brinner \\
   Computational Linguistics\\Department of Linguistics \\
  Bielefeld University, Germany \\
  {\tt marc.brinner@uni-bielefeld.de} \\\And
  Tarek Al Mustafa \\
  Heinz Nixdorf Chair for\\Distibuted Information Systems\\Institute of Computer Science \\
  Friedrich Schiller University Jena, Germany \\
  {\tt tarek.almustafa@uni-jena.de} \\\AND
  Sina Zarrieß \\
  Computational Linguistics\\Department of Linguistics \\
  Bielefeld University, Germany \\
  {\tt sina.zarriess@uni-bielefeld.de} \\}

\begin{document}
\maketitle

\begin{abstract}
We investigate the use of LLM-generated data for continual pretraining of transformer encoder models in specialized domains with limited training data, using the scientific domain of invasion biology as a case study. To this end, we leverage domain-specific ontologies by enriching them with LLM-generated data and pretraining the encoder model as an ontology-informed embedding model for concept definitions. To evaluate the effectiveness of this method, we compile a benchmark specifically designed for assessing model performance in invasion biology. After demonstrating substantial improvements over standard MLM pretraining, we investigate the feasibility of applying the proposed approach to domains without comprehensive ontologies by substituting ontological concepts with concepts automatically extracted from a small corpus of scientific abstracts and establishing relationships between concepts through distributional statistics. Our results demonstrate that this automated approach achieves comparable performance using only a small set of scientific abstracts, resulting in a fully automated pipeline for enhancing domain-specific understanding of small encoder models that is especially suited for application in low-resource settings and achieves performance comparable to masked language modeling pretraining on much larger datasets.

\end{abstract}

\section{Introduction}

Transformer encoder models such as BERT \cite{devlin2019bert} and its successors (e.g., \citealp{liu2019roberta}, \citealp{he2021debertav3}, \citealp{warner2024modernbert}) have consistently achieved state-of-the-art results across a wide range of NLP tasks. These successes are largely driven by large-scale pretraining on general-domain corpora such as Wikipedia and BookCorpus \cite{zhu2015books}, using objectives like masked language modeling (MLM) or replaced token detection \cite{clark2020electra}.

While transformer encoders offer an optimal balance between performance and efficiency, their full effectiveness in specialized domains - such as scientific text processing - is often enabled by additional pretraining on domain-specific corpora \cite{beltagy2019scibert, 9791256}, proven highly effective in fields where extensive domain-specific data is available (e.g., biomedical text processing \citealp{Gu_2021}). However, in specialized disciplines with limited resources, the potential of this approach diminishes, highlighting the need for alternative methods of injecting domain knowledge during pretraining.

To address this challenge, we propose a novel method for continual pretraining of transformer encoder models that leverages a set of domain-relevant concepts and their corresponding definitions as the core training resource. These concepts can be drawn from domain-specific ontologies (i.e., data structures containing precise, domain-specific and structured knowledge curated by domain experts \citet{walls2014semantics, 10.1093/sysbio/syad025, alsayed_algergawy_2025_14864244}) or extracted from texts using LLMs. Using this resource, we pretrain the model as an embedding model for concept definitions, encouraging definitions of identical or related concepts to occupy nearby positions in the embedding space and thus enabling the model to develop a structured understanding of domain-specific entities and their interconnections.


In this process, we perform an extensive analysis of the effectiveness of incorporating different types of information, moving from using domain-relevant concepts extracted from the ontologies towards a fully unsupervised pipeline with LLM-extracted concepts.

\noindent\textbf{Contributions:}

1) We validate the effectiveness of our embedding-based pretraining approach using ontology-derived concepts and LLM-generated definitions, establishing it as a viable alternative to traditional MLM pretraining.

2) We identify the benefits of incorporating concept relatedness by integrating ontological relationship links into the pretraining objective.

3) We explore the possibility of combining our pretraining strategy with traditional MLM pretraining, demonstrating strong synergistic effects that vastly improve downstream performance.

4) We create and evaluate a fully unsupervised pipeline by replacing ontology-derived concepts with LLM-extracted concepts from scientific abstracts. By also using distributional statistics as a concept relatedness indicator, we remove the dependency on manually curated ontologies.

5) We analyze performance of our unsupervised approach across varying dataset sizes, showing that it consistently outperforms MLM pretraining, even when trained on significantly less data.

6) We focus on the use of synthetic data in the form of LLM-generated concept definitions and analyze model collapse \cite{shumailov2024curserecursiontraininggenerated}, demonstrating that our proposed pretraining scheme is much less susceptible to this issue compared to classical mask language modeling.



Due to the extensiveness of our multi-step experiments and analysis, we focus on a single, representative domain: invasion biology. This field exemplifies a complex, specialized area of scientific research with limited unsupervised pretraining data or annotated resources. To support evaluation, we compile a new benchmark from three existing studies \cite{brinner2022linking, brinn2024weaklyclaim, brinner2025efficient}, covering a diverse set of tasks that collectively provide a comprehensive assessment of model performance in this domain.

\section{Related Work}
\label{sec:related_work}

\noindent\textbf{Continual Pretraining} is an effective and efficient approach to make LMs robust against new, ever-changing data that differs from its original pretraining \cite{ wu2024continuallearninglargelanguage, zhou2024continuallearningpretrainedmodels, parmar2024reusedontretrainrecipe, shi2024continuallearninglargelanguage}, enhances an LLM's domain specific effectiveness \cite{gururangan2020dontstoppretrainingadapt, gong-etal-2022-continual, xie2023efficientcontinualpretrainingbuilding, yıldız2025investigatingcontinualpretraininglarge} and improves knowledge transfer to downstream tasks \cite{10444954}.

\noindent\textbf{Ontologies and Knowledge Graphs} (KGs) have been explored as resource for continual pretraining since they provide a structured representation of domain knowledge in the form of unique entities and precise relations between them, contrasting the distributed and often less precise knowledge representation within neural networks. To bridge this gap, various methods have been proposed to integrate structured knowledge into transformer models. While some approaches incorporate KG information during inference \cite{zhang2019ernie, peters2019knowledge, he2020bert}, the majority of approaches focus on creating KG-informed pretraining methods, for example by performing MLM pretraining that incorporates knowledge about entities \cite{shen2020exploiting, zhang2021ebert}, performing MLM pretraining on sentences derived from KG triples \cite{lauscher2020common, moiseev2022skill, liu2022enhancing, sahil2023leveraging, Omeliyanenko2024preadapter}, designing auxiliary classification tasks based on ontological knowledge \cite{wang2021k, glauer2023ontology} or by creating contrastive ontology-informed sentence embedding methods \cite{wang2021kepler, ronzano2024ontology}. Our approach aligns most closely with the latter but extends it into a broader framework that incorporates not only relationships between concepts but also LLM-derived knowledge about individual concepts by incorporating synthetic concept definitions, thus creating a more informative and flexible pretraining process that is not reliant on the presence of ontologies.

\noindent\textbf{Using Synthetic Data} for model pretraining and/or fine-tuning is an appealing prospect \cite{long2024llmsdrivensyntheticdatageneration}, especially in specialized domains with little available training data.
Many studies explore the potential of LLM-generated or LLM-annotated data to enhance task-specific performance, both for encoder \cite{kruschwitz-schmidhuber-2024-llm, kuo2024llmgenerateddataequalrethinking, wagner2024powerllmgeneratedsyntheticdata} and decoder architectures \cite{ren2024learn, lee2024llm2llmboostingllmsnovel}. 

Beyond task-specific fine-tuning, synthetic data has also been investigated for task-agnostic pretraining. While this approach has shown promise for general-domain models \cite{alcoba2024utility, yang2024syntheticcontinuedpretraining, 10.1007/978-3-031-73397-0_18}, its application in domain-specific pretraining remains relatively underexplored (e.g., \citealp{yuan2024continuedpretrainedllmapproach}).




Despite its advantages, synthetic data introduces risks, including potential performance degradation compared to human-generated data - a phenomenon known as \textit{model collapse} \cite{shumailov2024curserecursiontraininggenerated}, prompting studies aimed at mitigating this effect, especially for autoregressive LLMs \cite{bertrand2024stabilityiterativeretraininggenerative, gerstgrasser2024modelcollapseinevitablebreaking, zhang2024regurgitativetrainingvaluereal, zhu2024synthesizetextdatamodel}. In Section \ref{sec:discussion}, this phenomenon will be further discussed in the context of our own experiments.


\section{Method}
\label{sec:method}

We propose a method for injecting domain knowledge into transformer models through continual pretraining. This section provides a general overview of our approach, while Section \ref{sec:experiments_onto} and Section \ref{sec:experiments_abstract} detail and evaluate its application to datasets derived from ontologies and scientific abstracts.

\subsection{Similarity-Based Pretraining}

We propose a pretraining strategy for transformer encoder models, training them as embedding models for concept definitions by teaching it to place definitions of the same concept or definitions of related concepts to similar positions in the embedding space, thus enabling the model to capture both the meaning and distinctions between domain-specific concepts effectively. A comparable strategy has previously proven effective for training specialized embedding models on scientific abstracts, where it substantially improved semantic encoding capabilities \cite{brinner2025semcse}, thus suggesting that a related approach may provide an effective means of enforcing semantic understanding of relevant domain knowledge.

Our method operates on a dataset of domain-relevant concepts $\mathcal{C}  = \{C_1, C_2, ...\}$, each in combination with multiple natural language concept definitions $\mathcal{D} = \{(d_{1, 1}, d_{1, 2}, ...), (d_{2, 1}, d_{2, 2}, ...), ...\}$.
Also, we optionally incorporate a set of tuples indicating pairs of related concepts $\mathcal{R} = \{(C_i, C_j), ...\}$ to increase the model's domain understanding beyond knowledge of individual entities.

The core training scheme is as follows: Given two concepts $C_i$ and $C_j$ from the dataset, we train the model to embed two definitions of concept $C_i$ to nearby locations in the embedding space while positioning a definition of $C_j$ further away, thereby teaching the model an understanding of the different concepts. This is achieved by sampling two definitions $d_{i, 1}$ and $d_{i, 2}$ that define concept $C_i$, and one definition $d_{j, 1}$ that defines concept $C_j$. These definitions are then mapped into the high-dimensional embedding space using the model $M$, resulting in embeddings $e_{i, 1}$, $e_{i, 2}$ and $e_{j, 1}$.

In practice, the embedding corresponds to the model’s output vector at the CLS token. To encourage the model to map definitions of the same concept in the embedding space to similar locations, we employ a triplet margin loss:
\begin{align*}
L = \textrm{relu}(||e_{i, 1} - e_{i, 2}|| - || e_{i, 1} - e_{j, 1}|| + 1)
\end{align*}
In this triplet loss formulation, $d_{i, 1}$ serves as an \textit{anchor}, with $d_{i, 2}$ being the \textit{positive} and $d_{j, 1}$ being the \textit{negative} with respect to that anchor. The loss function thus penalizes cases in which the distance between the anchor and the positive (i.e., two definitions defining the same concept) is not at least one unit (a margin hyperparameter) smaller than the distance between the anchor and the negative.

Rather than explicitly sampling individual triplets (anchor, positive, and negative), we optimize the loss computation by leveraging in-batch negatives, thus only sampling an anchor and a positive for each concept and using all definitions from other concepts within the batch as negatives. This strategy - in combination with switching the roles of anchor and positive - significantly increases the number of triplets contributing to the loss, leading to $4\cdot (n-1)$ triples per anchor-positive pair with a batch-size of $n$. This substantial increase in triplets enhances model performance, as the loss function quickly reaches zero for many triplets after just a few epochs due to the model's rapidly improving embedding capabilities. Consequently, the larger number of triplets increases the likelihood of encountering more informative gradient signals, ultimately leading to more effective embeddings.

\subsection{Concept Relatedness}

The current loss formulation encourages the model to map similar definitions (i.e., those defining the same concept) to nearby positions in the embedding space. While this enhances the model’s ability to differentiate between concepts, a deeper understanding of the domain also requires learning relationships between different concepts, which might otherwise be learned only implicitly through the similarity between their definitions. Therefore, we extend our loss formulation by incorporating additional triplets that capture concept relatedness.

Specifically, if two concepts $C_i$ and $C_j$ are in the same batch and $(C_i, C_j) \in \mathcal{R}$, we treat their definitions as additional positive pairs within the loss function, while using the definitions of all unrelated concepts as negatives. This setup implicitly introduces a ranking effect: definitions of the same concept are drawn closest together, as the corresponding loss triples include all other definitions - related or not - as negatives. In contrast, triples based on related concepts use only definitions of completely unrelated concepts as negatives, thereby encouraging related concepts to be embedded closer to one another than unrelated ones.

\subsection{Pretraining Loss Combination}
Our proposed loss is applied to the CLS token representation, allowing seamless integration with other pretraining losses that target the remaining token embeddings. This is especially interesting in light of recent models being trained exclusively with MLM loss \cite{warner2024modernbert}, since the traditional next sentence prediction loss empirically did not lead to significant performance gains \cite{liu2019roberta}.
Consequently, our method presents a more sophisticated approach of infusing domain-relevant knowledge into the CLS token representation.

\section{Ontology-Informed Pretraining}
\label{sec:experiments_onto}

This section details the application and evaluation of our proposed method, using domain-specific ontologies for dataset creation. Our experiments focus on the scientific domain of invasion biology, a specialized subfield of biodiversity research that investigates non-native species, their introduction pathways, ecological impacts, and management strategies to mitigate their effects on ecosystems \cite{jeschke2018invasion}.

\subsection{Dataset Creation}
\label{sec:onto_data_creation}

Our approach involves constructing a domain-specific dataset consisting of concepts, definitions and concept relations in the target domain. To this end, we use two ontologies that address the target domain: the INBIO ontology \cite{alsayed_algergawy_2025_14864244}, which captures concepts relevant to invasion biology, and the ENVO ontology \cite{buttigieg2013environment, buttigieg2016environment}, which provides a structured representation of environmental and ecological concepts.

From these ontologies, we extract concept-definition pairs for all concepts that have a corresponding definition, as well as relational links between concepts. Additionally, we use a LLM, LLaMA-3-8B-Instruct \cite{grattafiori2024llama3herdmodels}, to generate five additional definitions per concept, using the original ontology definition as context during generation to ensure that the new definitions accurately reflect the domain-specific meaning.

We compare our proposed pretraining approach to traditional MLM pretraining on sentences extracted from scientific abstracts. 
We leverage an existing index of paper titles in the field of invasion biology \cite{Mietchen2024ni} and employ a web scraper to retrieve their abstracts, resulting in a final collection of 37,786 paper titles and abstracts.

Since we explicitly aim to assess the applicability of our approach in low-resource settings, most experiments are conducted on a subset of 5,000 abstracts. This results in a dataset containing 47,031 sentences extracted from 5,000 scientific abstracts, alongside 5,197 ontology-derived concepts, each supplemented with at least one extracted definition and five generated definitions.

\subsection{Model Pretraining}

In our experiments, we perform continual pretraining on a DeBERTa-base model \cite{he2021deberta} by leveraging three different pretraining strategies:

\begin{enumerate}
    \item \textbf{Masked language modeling (MLM) pretraining} with a masking probability of 0.25, applied to either abstract sentences, generated definitions, or a combined dataset of both.
    \item \textbf{Similarity (SIM) pretraining} as described in Section \ref{sec:method}, using our proposed similarity-based approach on the ontology-derived data.
    \item \textbf{Combined pretraining} using MLM and SIM losses, done to investigate potential synergies between these approaches. We apply both strategies concurrently by performing two forward and backward passes - one for each loss function - for each parameter update.
\end{enumerate}

Further details about the pretraining can be found in Appendix \ref{sec:appendix_pretraining}.

\subsection{Evaluation Datasets}

Building on existing studies, we compile a benchmark comprising four distinct tasks in invasion biology, each with unique evaluation requirements.

The \textbf{Hypothesis Classification} task \cite{brinner2022linking} is a 10-class classification task on identifying which of 10 major hypotheses in invasion biology is addressed in a given scientific abstract. Due to class imbalance, we report both micro F1 and macro F1 scores.

The \textbf{Hypothesis Span Prediction} task \cite{brinn2024weaklyclaim} is a token-level prediction task based on the same abstracts as the INAS classification task. Annotators provide span-level evidence annotations for each hypothesis and we evaluate the model's ability to predict the tokens that were annotated (Token F1) as well as the ability to recognize complete spans (Span F1).

The \textbf{EICAT Impact Classification} task \cite{brinner2025efficient} is a classification task on assessing the impact of an invasive species as reported in a given scientific full text, assigning it to one of six predefined impact categories. We evaluate performance using macro F1 and micro F1 scores.

The \textbf{EICAT Impact Evidence} task \cite{brinner2025efficient} leverages evidence annotations provided by the EICAT classification dataset, created by domain experts who identified sentences in the full-texts indicating the species' impact category. We evaluate the model’s ability to rank relevant sentences within a full text using the normalized discounted cumulative gain (NDCG) metric.

\begin{table*}[h]
\setlength\tabcolsep{4pt}
  \centering
  \begin{tabular}{l|cc|cc|cc|c|c}
    \hline
     \multicolumn{1}{c|}{} & \multicolumn{2}{c|}{\small \bf Hypothesis Clf} & \multicolumn{2}{c|}{\small \bf Hypothesis Span} & \multicolumn{2}{c|}{\small \bf Impact Clf} & \multicolumn{1}{c|}{\small \bf Impact Evid.} & \multicolumn{1}{c}{\small \bf Avg.}\\
    \textbf{Model} & \small Macro F1 & \small Micro F1 & \small Token F1 & \small Span F1 & \small Macro F1 & \small Micro F1 & \small NDCG \\
    \hline
    \small DeBERTa base & 0.674 & 0.745 & 0.406 & 0.218 & 0.392 & 0.416 & 0.505 & 0.483 \\ \hline
    \multicolumn{9}{c}{\bf MLM Pretraining} \\ \hline
    \small Abstract Sentences & 0.744 & 0.792 & 0.413 & 0.219 & 0.433 & 0.455 & 0.499 & 0.507 \\
    \small Ontology Definitions & 0.685 & 0.759 & 0.409 & 0.222 & 0.448 & 0.446 & 0.501 & 0.496 \\
    \small Keyword Definitions & 0.719 & 0.776 & 0.397 & 0.194 & 0.428 & 0.441 & 0.478 & 0.492 \\
    \small Abstract Sent.+Ontology Def. & 0.740 & 0.804 & 0.415 & 0.230 & 0.459 & 0.479 & 0.512 & 0.519 \\
    \small Abstract Sent.+Keyword Def. & 0.729 & 0.799 & 0.417 & 0.221 & 0.439 & 0.455 & 0.497 & 0.507 \\ \hline
    \multicolumn{9}{c}{\bf Similarity Pretraining} \\ \hline
    \small Ontology Definitions & 0.727 & 0.779 & 0.400 & 0.218 & 0.446 & 0.460 & 0.514 & 0.507 \\
    \small Keyword Definitions & 0.726 & 0.783 & 0.405 & 0.228 & 0.465 & 0.475 & 0.497 & 0.510 \\ \hline
    \multicolumn{9}{c}{\bf MLM+Similarity Pretraining} \\ \hline
    \small Abstract Sent.+Ontology Def. & 0.750 & 0.812 & 0.414 & 0.242 & 0.504 & 0.518 & 0.530 & 0.538 \\
    \small Abstract Sent.+Keyword Def. & 0.740 & 0.805 & 0.415 & 0.220 & 0.469 & 0.489 & 0.511 & 0.520 \\ \hline
    \multicolumn{9}{c}{\bf Other Domain-Specific Models} \\ \hline
    \small PubMedBERT & 0.728 & 0.783 & 0.410 & 0.208 & 0.509 & 0.508 & 0.552 & 0.531 \\
    \small SciDeBERTa & 0.736 & 0.805 & 0.417 & 0.213 & 0.468 & 0.484 & 0.494 & 0.514 \\ \hline
  \end{tabular}
  \caption{Benchmark results for different pretraining methods leveraging either the ontology or a dataset of 5000 scientific abstracts, as well as a comparison to two pretrained models from the biomedical domain.}
  \label{tab:results1}
  \vspace{-10px}
\end{table*}

These tasks address different aspects of the field of invasion biology but have in common that they require extensive domain knowledge for a deep interpretation of scientific texts within the broader context of the field. Taking the hypothesis classification tasks as an example, this could manifest itself in needing to identify a hypothesis solely by means of a description of an experimental design or measurements taken within an ecosystem.

To mitigate variance inherent to model training, we train 7 models for the hypothesis and impact classification tasks and 3 models for the remaining tasks and report the average performance. For details on task setup, dataset sizes and training methodologies, please refer to Appendix \ref{sec:appendix}.

To obtain a single benchmark score, we compute task-specific scores by averaging the individual performance metrics for each task and averaging the results across all four tasks.

\subsection{Results}

The results of our evaluation of different pretraining methods are presented in Table \ref{tab:results1}.

First, we observe that traditional MLM pretraining on sentences extracted from just 5,000 scientific abstracts yields significant improvements across all tasks compared to the DeBERTa baseline, raising the benchmark score from 0.483 to 0.507.

As a baseline, we also assess the impact of MLM pretraining on synthetic definitions. While this also resulted in increased performance, the gains are smaller than those achieved through pretraining on abstract sentences. Additionally, despite the datasets being of similar size, optimal performance with synthetic definitions is reached after approximately 40K batches, in contrast to 200K batches for MLM on abstract sentences, which is analyzed further in Section \ref{sec:discussion_collapse}.

As a last MLM baseline, we investigate MLM pretraining on a mixture of synthetic definitions and abstract sentences. Since initial experiments using a 1:1 ratio led to worse results compared to training on abstract sentences alone, we adjusted the ratio to 1:3 (ontology definitions to abstract sentences), resulting in improved performance compared to using abstract sentences alone and suggesting that concept definitions provide useful additional information to the model.

Turning to our proposed embedding similarity (SIM) pretraining approach, we find that applying it to ontology definitions achieves performance on par with MLM pretraining on real data (both scoring 0.507), establishing our method as viable alternative in the absence of such data. However, since SIM pretraining only affects the CLS token representation, we observe (on average) increased performance on classification tasks while performance decreased on the token-level prediction task, indicating that our approach primarily enhances the representation of the entire input sequence.

The most notable improvements arise when combining SIM pretraining on synthetic ontology definitions with MLM pretraining on abstract sentences. This approach leads to substantial performance gains across most tasks compared to MLM pretraining alone. Specifically, the overall benchmark score increases from 0.507 (MLM on abstract sentences) to 0.538. Notably, the substantial improvement over using either pretraining method individually (or over using the combined data for MLM) suggests a synergistic effect, indicating that SIM pretraining enhances the understanding of individual concepts, while MLM pretraining strengthens the model’s grasp of relationships between concepts and general language understanding. As a result, this combined approach outperforms models trained on millions of abstracts from the broader biomedical domain, such as PubMedBERT \cite{Gu_2021} and SciDeBERTa \cite{9791256}, which generally are strong baselines in this field \cite{brinner2022linking}.

Finally, we perform an ablation experiment by performing SIM pretraining without leveraging concept relatedness information. This leads to a significant drop in performance (0.498 compared to 0.507 with concept relatedness), suggesting that the relatedness encoded in ontologies is a useful training signal (Appendix \ref{sec:appendix_ablation}, Table \ref{tab:ablation}).

\begin{table*}
  \centering
  \begin{tabular}{l|cc|cc|cc|c|c}
    \hline
     \multicolumn{1}{c|}{} & \multicolumn{2}{c|}{\small \bf Hypothesis Clf} & \multicolumn{2}{c|}{\small \bf Hypothesis Span} & \multicolumn{2}{c|}{\small \bf Impact Clf} & \multicolumn{1}{c|}{\small \bf Impact Evid.} & \multicolumn{1}{c}{\small \bf Avg.}\\
    \textbf{Model} & \small Macro F1 & \small Micro F1 & \small Token F1 & \small Span F1 & \small Macro F1 & \small Micro F1 & \small NDCG \\
    \hline
    \multicolumn{8}{c}{\bf MLM Pretraining} \\ \hline
    \small 5000 Abstracts & 0.744 & 0.792 & 0.413 & 0.219 & 0.433 & 0.455 & 0.499 & 0.507 \\
    \small 15000 Abstracts & 0.731 & 0.801 & 0.415 & 0.234 & 0.480 & 0.499 & 0.493 & 0.518 \\
    \small 25000 Abstracts & 0.748 & 0.807 & 0.418 & 0.233 & 0.460 & 0.484 & 0.512 & 0.522 \\
    \small 35000 Abstracts & 0.735 & 0.811 & 0.419 & 0.244 & 0.483 & 0.484 & 0.494 & 0.521 \\ \hline
    \multicolumn{9}{r}{\bf Avg: 0.517} \\ \hline
    \multicolumn{9}{c}{\bf MLM+Similarity Pretraining} \\ \hline
    \small 5000 Abstracts & 0.740 & 0.805 & 0.415 & 0.220 & 0.469 & 0.489 & 0.511 & 0.520 \\
    \small 15000 Abstracts & 0.754 & 0.812 & 0.418 & 0.245 & 0.474 & 0.489 & 0.519 & 0.529 \\
    \small 25000 Abstracts & 0.759 & 0.806 & 0.419 & 0.236 & 0.479 & 0.499 & 0.511 & 0.528 \\
    \small 35000 Abstracts & 0.756 & 0.824 & 0.418 & 0.241 & 0.477 & 0.489 & 0.551 & 0.538 \\ \hline
    \multicolumn{9}{r}{\bf Avg: 0.529} \\ \hline
  \end{tabular}
  \caption{Comparing MLM and combined MLM+SIM pretraining with keyword definitions for varying dataset sizes.}
  \label{tab:results2}
  \vspace{-10px}
\end{table*}

\section{Using LLM-Extracted Keywords}
\label{sec:experiments_abstract}

In the previous section, we explored the performance improvements achieved by combining our proposed similarity loss on ontology-derived data with traditional MLM pretraining. While this approach is highly valuable in domains with available ontologies, many fields may lack such structured resources. To address this limitation, we explore the feasibility of using an LLM for constructing a dataset of domain-relevant concepts, definitions, and relations using only a small set of scientific abstracts. We compare results achieved on our original dataset of 5,000 abstracts with those using ontology-derived data and also evaluate how well our approach scales with increasing dataset size.

\subsection{Dataset Creation}
\label{sec:abstract_data_creation}

To construct the dataset, we assume access to a small collection of scientific abstracts, as discussed in Section \ref{sec:onto_data_creation}.The dataset $(\mathcal{C}, \mathcal{D}, \mathcal{R})$ is obtained through the following three steps:

\begin{enumerate}
    \item \textbf{Keyword Extraction}: We extract domain-relevant concepts in the form of keywords from scientific abstracts using LLaMA-3-8B \cite{grattafiori2024llama3herdmodels}. This is achieved by appending the string "Keywords:" to each abstract and allowing the language model to generate a continuation, effectively identifying key concepts within the text.
    \item \textbf{Definition Generation}: For each extracted keyword, we generate five additional definitions using LLaMA-3-8B-Instruct. To ensure that the generated definitions accurately reflect domain-specific usage, the original abstract from which the keyword was extracted serves as context during generation.
    \item \textbf{Relation Identification}: We determine concept relationships by analyzing co-occurrence patterns within the abstracts. Keyword names are first normalized using stemming, followed by exact string matching to identify equivalent keywords across different abstracts. Two keywords are considered related if they co-occur more than $k$ times (a tunable hyperparameter), with all other samples serving as negatives.
\end{enumerate}

We again begin by evaluating results on a dataset of 5,000 abstracts, which constrains both the number of abstract sentences available for pretraining as well as the number of extracted keywords with corresponding definitions created within our pipeline, resulting in 23,597 unique keywords. This setup allows us to assess the effectiveness of our approach in a low-resource setting. We then examine the impact of dataset size by progressively increasing the number of abstracts to 15,000, 25,000, and 35,000.

\subsection{Results}

Results for the first set of experiments operating on 5000 scientific abstracts are displayed in Table \ref{tab:results1}.

We again evaluate MLM pretraining on the new dataset of LLM-generated keyword definitions as a baseline, which leads to slight improvements over the standard DeBERTa base model by achieving scores of 0.492 when trained solely on keyword definitions and 0.507 when combined with abstract sentences. However, these gains are less pronounced than those using LLM-generated definitions for ontological concepts, indicating that ontological concepts offer more valuable information to the encoder model (compare Section \ref{sec:discussion}).

In contrast, SIM pretraining on keyword definitions yields slightly better performance than using ontology definitions, which may be attributed to dataset size as the LLM extracted 23,597 unique keywords from the abstracts, compared to 5,179 concepts from the ontologies. Notably, this lets SIM pretraining on data extracted from 5,000 abstracts outperform MLM pretraining on that same dataset, thus validating our proposed pretraining approach and suggesting that the LLM has enriched our base dataset with valuable information.

Combining SIM and MLM pretraining again leads to improved results compared to either strategy alone, thus undermining the synergistic effects. However, the performance gains are weaker than those achieved using the ontology-derived data (0.520 vs. 0.538), which we analyze further in Section \ref{sec:discussion}. Still, the resulting model using just 5,000 abstracts outperforms SciDeBERTa, which was trained on millions of scientific abstracts.

Lastly, we assess the effect of varying dataset sizes on our pretraining pipeline. While an increase in data availability leads to more detected keywords for SIM pretraining, it also leads to more abstract sentences for MLM pretraining, which may diminish the relative value added by the LLM. However, as shown in Table \ref{tab:results2}, even with larger datasets, our fully automated knowledge injection strategy consistently outperforms traditional MLM pretraining, even though both are based on the same dataset.

Despite efforts to mitigate variance by training multiple models per task, individual results still remain subject to fluctuation (see Appendix \ref{sec:appendix_significance} for an analysis on statistical significance). Therefore, we consider the average scores across all dataset sizes - 0.517 for MLM pretraining and 0.529 for combined pretraining - as the most reliable indicators of the substantial performance improvements achievable with our pipeline.

\section{Discussion}
\label{sec:discussion}

\subsection{Are Ontologies Replaceable?}
Our experiments demonstrate that injecting domain-specific knowledge from ontologies into encoder models can substantially enhance downstream performance. Notably, we also found that this knowledge can - to some extent - be replaced by a combination of LLM-extracted keywords, definitions, and co-occurrence statistics. Still, we argue that ontologies are a more valuable resource, which is supported by several observations.

First, despite our automated pipeline extracting a significantly larger number of keywords from 5,000 abstracts than were present in the ontologies (23,597 vs. 5,179), MLM pretraining performance was better using ontology-based data. This suggests that ontology-derived data is of higher quality, likely due to the careful selection of domain-relevant concepts, making even small ontologies highly valuable. In contrast, many automatically extracted keywords, such as species names, may be less informative for analyzing species invasions than more targeted ontology concepts.

Second, we  find that a combination of synthetic data and abstract sentences leads to superior results when ontology-based definitions are used instead of keyword definitions (both for MLM and SIM). This disparity may stem from the fact that information extracted from the abstracts is inherently tied to the same dataset, thus offering less additional insight compared to the disconnected and therefore more informative ontology.

Finally, ontological relations encode different knowledge compared to statistical co-occurrence patterns. Most relations within the investigated ontologies were subclass relations, that contribute to a refined hierarchical understanding of domain-specific concepts. In contrast, co-occurrence statistics primarily capture broader associations between concepts within the domain and the contexts they appear in. Our results indicate that both types of information benefit model pretraining, but we do not believe that they should be equated.

\subsection{Investigating Model Collapse}
\label{sec:discussion_collapse}

Previous studies have identified a risk of model collapse when training on LLM-generated data (see Section \ref{sec:related_work}). Similarly, we observed that both MLM and SIM training on synthetic data reached peak performance after approximately 40K batches, after which performance began to decline. In contrast, training on the dataset consisting of abstract sentences peaked at around 200K batches, with performance remaining stable even when training for twice as long. This suggests that while the generated data provides valuable information, excessive use can still lead to model collapse.

It is important to note that we cannot conclusively attribute this behavior solely to the synthetic nature of the data. Since the generated dataset consists exclusively of concept definitions, its inherently lower variance compared to abstract sentences may contribute to catastrophic forgetting of broader language understanding, rather than model collapse in the strict sense.

Nevertheless, we found that performance degradation with synthetic data was much less pronounced for SIM training compared to MLM. This is likely due to weaker gradient signals after the peak has been reached, as most training triples eventually reach zero loss. This has the positive effect that, when SIM pretraining on synthetic data is combined with MLM training on abstract sentences, the risk of model collapse is effectively mitigated because the weak (but still informative) gradients from SIM training are not strong enough to induce this effect. 

This is in contrast to MLM training on a combination of abstract sentences and synthetic definitions. Here, performance declined compared to training on abstract sentences alone when both sources of data were used in equal proportion. This suggests that in this setting, the signal leading to model collapse is too strong, leading us to adopt a 1:3 ration in our experiments.

Ultimately, these findings highlight the advantage of our proposed pretraining scheme over traditional MLM, as it enables effective utilization of synthetic data while avoiding detrimental effects on model stability.

\section{Conclusion}
In this study, we investigated the use of LLM-generated, synthetic data for continual pretraining of domain-specific encoder models, demonstrating how to utilize domain specific ontologies or derive domain information through LLM-extraction from scientific abstracts for domains where ontologies may not be available.

Our results demonstrate that the proposed pretraining approach produces strong synergistic effects when combined with masked language modeling training. This leads to significant performance improvements in low-resource settings and results in a model surpassing other specialized models from the broader biomedical domain, despite being trained on orders of magnitude less data.



Given the minimal data requirements, our approach has the potential to be widely applicable beyond the domain explored in this study. Furthermore, its robustness against model collapse despite using synthetic data represents a meaningful advancement in leveraging LLM-generated data for training specialized models.

\section{Limitations}

We note several limitations of our approach:
First, while we demonstrate strong performance in the domain of invasion biology, its applicability to other domains remains uncertain and requires further evaluation, which was not possible to include within this study given the extend of the existing evaluations and analyses.

Second, although we compare the effectiveness of leveraging information from an ontology versus extracting it from scientific abstracts, our comparison is constrained by the specific ontology elements considered - namely, the selection of concepts, their definitions, and the presence of links. We believe that significant untapped potential remains in additional ontology features, such as relation types, domains and ranges of relations, and higher-order relationships. A more comprehensive assessment of the ontology’s value can only be made once its full informational capacity is utilized.

Third, assessing the correctness and quality of LLM-generated data and extracted concepts from scientific abstracts is beyond the scope of this study. While our results indicate performance improvements on the invasion biology benchmark, there remains a risk of introducing bias or inaccuracies into the encoder model due to biased concept selection or potential misinterpretations by the LLM.

\section{Acknowledgments}

This work was funded by the European Regional Development Fund within the project "LLM4KMU - Optimierter Einsatz von Open Source Large Language Modellen in KMU", and was partially produced within the focus group "Mapping Evidence to Theory in Ecology: Addressing the Challenges of Generalization and Causality" at the Center for Interdisciplinary Research, Bielefeld.

\bibliography{custom}

\appendix

\section{Experimental Details}
\label{sec:appendix}

Code for training and evaluation, training datasets and the best-performing encoder model checkpoint are available at \href{https://https://github.com/inas-argumentation/Ontology_Pretraining}{github.com/inas-argumentation/Ontology\_Pretraining}.

\begin{figure*}[h]
\fbox{
\begin{minipage}{0.99\textwidth}
\small{
Task: Create a single sentence that defines the concept listed below. You also receive an existing definition of the concept.\vspace{1em}

If you feel like the definition does not contain enough information, please create a more extensive one. If you feel like all necessary information is already contained, you do not need to add additional information. Please do not simply repeat the definition given to you. Please do not use the term itself in the definition.\vspace{1em}

Concept: [CONCEPT NAME]

Definition: [CONCEPT DEFINITION]\vspace{1em}

Format your response as:

Definition: [New Definition]

END.
}
\end{minipage}}
\caption{The Llama-3-8B-Instruct prompt for generating alternative definitions for concepts from the ontology.}
\label{fig:prompt_onto}
\vspace{-10px}
\end{figure*}

\subsection{Data Generation}
\label{sec:appendix_data_generation}
We used LLMs, specifically LLaMA-3-8B and LLaMA-3-8B-Instruct, to generate synthetic data for pretraining the encoder model. For generating alternative definitions of ontology concepts, we employed the instruction-tuned version of LLaMA, using the prompt shown in Figure \ref{fig:prompt_onto}.

Concepts were extracted from scientific abstracts following the procedure detailed in Section \ref{sec:abstract_data_creation}. Definition generation was then performed using a similar prompting approach, incorporating the scientific abstract as context.

Concept relations are identified using co-occurrence counts as described in Section \ref{sec:abstract_data_creation}. For the dataset consisting of 5000 abstracts, we treat concepts as related if they co-occur in at least 5 abstracts, which we selected manually by observing and assessing exemplary related concepts. Since many concepts occur rarely, this lead to each concept being on average related to about 0.5 other concepts. For larger dataset sizes, we adjust the number of co-occurrences that are required for two concepts to be related so that the number of related concepts for each concept stays roughly constant at 0.5, thus leading to a comparable assessment.

\subsection{Model Training}
\label{sec:appendix_pretraining}

We evaluate various pretraining strategies. Initially, we selected the optimal model checkpoint based on validation loss; however, we found that training for significantly longer improved downstream performance, even when the validation loss did not decrease. For this reason, we adopted a strategy of saving model checkpoints at different epochs and evaluating them on the INAS classification task. We then used this evaluation to identify the number of batches that are optimal for a given pretraining method. Once this number is established, we retrained the final models used in our evaluation from scratch using the predetermined number of epochs.

For similarity-based pretraining, we adopt a sampling strategy that increases the likelihood of samples that are related to each other being included within the same batch.

In the case of combined SIM and MLM pretraining, we independently sample a batch for each pretraining method and perform two backward passes - one for each loss - before applying a single parameter update.

For MLM pretraining, we found that a high weight decay value of 1e-2 was beneficial, likely mitigating overfitting to the small dataset. In contrast, for SIM pretraining we did not use weight decay, since applying it led to reduced downstream performance, potentially due to accelerated catastrophic forgetting of the model’s general language modeling capabilities if no MLM loss is used.

For combined pretraining, we again applied a weight decay of 1e-2.

\subsection{Evaluation Dataset}

\subsubsection{INAS Classification}

The INAS classification task \cite{brinner2022linking} is a 10-class classification problem, where the goal is to determine which of 10 prominent hypotheses are addressed in a given scientific abstract. We use the updated labels provided by \cite{brinn2024weaklyclaim}. The task is a multi-label classification task, meaning that multiple hypotheses can be addressed within a single abstract.

The dataset consists of 954 samples, with 721 used for training, 92 for validation, and 141 for testing. Models are trained as standard classifiers with a sigmoid activation function and a weighted binary cross-entropy loss. Given the highly imbalanced nature of the dataset, we report both micro and macro F1 scores to assess overall predictive performance as well as the ability to recognize underrepresented classes. Further details are available in our code repository.

\begin{table*}[ht!]
\setlength\tabcolsep{4pt}
  \centering
  \begin{tabular}{l|cc|cc|cc|c|c}
    \hline
     \multicolumn{1}{c|}{} & \multicolumn{2}{c|}{\small \bf Hypothesis Clf} & \multicolumn{2}{c|}{\small \bf Hypothesis Span} & \multicolumn{2}{c|}{\small \bf Impact Clf} & \multicolumn{1}{c|}{\small \bf Impact Evid.} & \multicolumn{1}{c}{\small \bf Avg.}\\ \hline
    \multicolumn{9}{c}{\bf Similarity Pretraining} \\ \hline
    \small Ontology Definitions & 0.727 & 0.779 & 0.400 & 0.218 & 0.446 & 0.460 & 0.514 & 0.507 \\
    \small Keyword Definitions & 0.726 & 0.783 & 0.405 & 0.228 & 0.465 & 0.475 & 0.497 & 0.510 \\ \hline
    \multicolumn{9}{c}{\bf Similarity Pretraining Ablation: No Concept Relatedness} \\ \hline
    \small Ontology Definitions & 0.715 & 0.777 & 0.395 & 0.210 & 0.436 & 0.450 & 0.499 & 0.498 \\
    \small Keyword Definitions & 0.725 & 0.781 & 0.402 & 0.209 & 0.466 & 0.484 & 0.484 & 0.504 \\ \hline
  \end{tabular}
  \caption{Results for an ablation study, evaluating the effect of not using the relatedness between different concepts in the pretraining loss.}
  \label{tab:ablation}
\end{table*}

\subsubsection{INAS Span Prediction}

The INAS Span Prediction task \cite{brinn2024weaklyclaim} is closely related to the INAS classification task and is based on the same dataset. However, instead of classifying abstracts, it involves identifying spans of text indicative of the 10 hypotheses, as annotated by human experts.

Only 750 samples contain token-level annotations. Models are trained using a weighted binary cross-entropy loss applied to 10 logits that were predicted for each input token, with each logit corresponding to one of the hypotheses. Additionally, we trained models as normal classifier as in the INAS classification task, where we also included all samples without token-level annotations.

We evaluate performance using two metrics:

\begin{itemize}
    \item \textbf{Token-F1 Score}: This score measures the ability to identify individual tokens as being indicative of a specific hypothesis (i.e., belonging to a ground-truth annotation).
    \item \textbf{Span-F1 Score}: This score evaluates how well models detect complete spans by assessing the intersection-over-union (IoU) between predicted and ground-truth spans at different thresholds.
\end{itemize}

For further details on these metrics, see \cite{brinn2024weaklyclaim}.

\subsubsection{EICAT Classification}

The EICAT classification task \cite{brinner2025efficient} is concerned with classifying the ecological impact of an invasive species as reported in a scientific full-text paper. The categories include five different impact levels plus a “Data Deficient” category, resulting in a six-class classification problem.

The dataset consists of 436 full-text scientific papers covering 120 species, with training, validation, and test splits of 82\%, 8\%, and 10\%, respectively.

Since most encoder models cannot process entire full-texts at once, \citet{brinner2025efficient} explored strategies for selecting relevant sentence subsets for training and evaluation. One effective and unbiased approach is the selection of random sentences, which we adopt. During testing, each model receives 20 different random sentence selections per paper, with the final classification determined via majority voting.

Models are trained as standard classifiers with a weighted categorical cross-entropy loss. Given the dataset’s class imbalance, we report both micro and macro F1 scores, following the approach used in the INAS classification task.

\subsubsection{EICAT Evidence Selection}

The EICAT evidence selection task \cite{brinner2025efficient} is a binary sentence classification problem. While annotating scientific full-texts for the EICAT classification task, human experts identified key sentences that served as evidence for impact assessments. The goal of this task is to predict whether a given sentence is evidence for an EICAT impact assessment.

To provide context, the model receives three sentences before and three sentences after the target sentence, with the target sentence enclosed by [SEP] tokens. Training is performed using a weighted binary cross-entropy loss.

The dataset splits are the same as those used in the EICAT classification task. Performance is reported using the normalized discounted cumulative gain (NDCG) score, which evaluates the model’s ability to rank ground-truth evidence sentences higher than non-evidence sentences. This metic is used since the task was proposed in the context of extracting a fixed number of sentences for further prediction, thus making the ranking between sentences more important than the specific predicted scores. Also, the original annotations are not guaranteed to include every sentence indicative of the correct classification, thus making a softer metric a better fit compared to a strict binary evaluation.

\subsection{Evaluation Details}

Due to the variance inherent to training models on evaluation tasks, we train 7 models for the INAS classification and EICAT classification tasks, as well as 3 models for the other tasks that take significantly longer for each training run. Final results are reported as the average performance across all runs. To compute a final benchmark score, we first average the performance metrics for each task separately and then compute an overall average across all tasks.

For some tasks, we observed occasional training runs (across all pretraining types) where models exhibited drastically lower performance caused by degenerate states that only predict a single class for all samples. We attribute this to the dataset’s extreme class imbalance, that, for some random seeds, leads to degenerate states that the model is unable to escape. In such cases, training runs were repeated to avoid reporting results that reflect random failures rather than actual model performance.

\subsection{Ablation}
\label{sec:appendix_ablation}
We perform an ablation study evaluating the effect of not incorporating the relations between different concepts (as determined by ontology relations or keyword co-occurrence statistics) into the pretraining loss. Results are displayed in Table \ref{tab:ablation}. We see that not incorporating concept relatedness leads to reduced scores on our benchmark, thus indicating the usefulness of leveraging this information within pretraining.

\subsection{Statistical Significance}
\label{sec:appendix_significance}

We perform multiple runs for each task to reduce variance in our reported results. To fully undermine our key results, we perform a permutation-based statistical significance test that takes all 20 (or 80 for the multiple dataset sizes) individual results that contribute to the final benchmark score into account. According to this, the following results are statistically significant ($p < 0.05$):
\begin{itemize}
    \item The superiority of all pretraining methods (except MLM pretraining using just keyword definitions) over DeBERTa base.
    \item The superiority of combining MLM pretraining on abstract sentences with similarity pretraining on ontology definitions compared to just MLM pretraining on abstract sentences.
    \item The superiority of combining MLM pretraining on abstract sentences with similarity pretraining on keyword definitions if evaluated over all dataset sizes.
    \item The superiority of MLM+SIM pretraining using ontology data over MLM+SIM pretraining using abstract-derived keyword data.
\end{itemize}

Thus, the following key insights are supported by statistical significance:
\begin{itemize}
    \item SIM pretraining alone is a valid pretraining strategy that improves performance.
    \item Combining SIM pretraining with MLM pretraining leads to improved results compared to just MLM pretraining alone. This holds both for the ontology-based and LLM-extracted keyword-based data.
    \item Ontology data is a more valuable resource than data reliant on LLM extracted keywords.
\end{itemize}
\end{document}